\documentclass{article}

\usepackage{arxiv}

\usepackage[utf8]{inputenc} % allow utf-8 input
\usepackage[T1]{fontenc}    % use 8-bit T1 fonts
\usepackage{hyperref}       % hyperlinks
\usepackage{url}            % simple URL typesetting
\usepackage{booktabs}       % professional-quality tables
\usepackage{amsfonts}       % blackboard math symbols
\usepackage{nicefrac}       % compact symbols for 1/2, etc.
\usepackage{microtype}      % microtypography
\usepackage{cleveref}       % smart cross-referencing
\usepackage{lipsum}         % Can be removed after putting your text content
\usepackage{graphicx}
\usepackage{natbib}
\usepackage{doi}
\usepackage[table]{xcolor}

\title{GEM: Empowering LLM for both Embedding Generation and Language Understanding}

\newif\ifuniqueAffiliation
\uniqueAffiliationtrue

\ifuniqueAffiliation % Standard variant of author block
\usepackage{authblk}

\newbox{\orcid}\sbox{\orcid}{\includegraphics[scale=0.06]{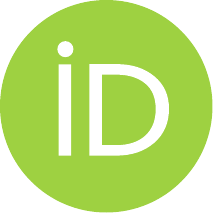}} 
\author[1]{Caojin Zhang}
\author[1*]{Qiang Zhang}
\author[1]{Ke Li}
\author[1]{Sai Vidyaranya Nuthalapati}
\author[1]{Benyu Zhang}
\author[1]{Jason Liu}
\author[1]{Serena Li}
\author[1]{Lizhu Zhang}
\author[1]{Xiangjun Fan}
\affil[1]{Meta Inc}
\fi

\hypersetup{
pdftitle={GEM: EMPOWERING LLM FOR BOTH EMBEDDING GENERATION
AND LANGUAGE UNDERSTANDING},
pdfsubject={q-bio.NC, q-bio.QM},
pdfauthor={},
pdfkeywords={LLM, embedding tasks, generation tasks},
}

\begin{document}
\maketitle

\begin{abstract}
Large decoder-only language models (LLMs) have achieved remarkable success in generation and reasoning tasks, where they generate text responses given instructions. However, many applications, e.g., retrieval augmented generation (RAG), still rely on separate embedding models to generate text embeddings, which can complicate the system and introduce discrepancies in understanding of the query between the embedding model and LLMs. To address this limitation, we propose a simple self-supervised approach, Generative Embedding large language Model (GEM), that enables any large decoder-only LLM to generate high-quality text embeddings while maintaining its original text generation and reasoning capabilities. Our method inserts new special token(s) into a text body, and generates summarization embedding of the text by manipulating the attention mask. This method could be easily integrated into post-training or fine tuning stages of any existing LLMs. We demonstrate the effectiveness of our approach by applying it to two popular LLM families, ranging from 1B to 8B parameters, and evaluating the transformed models on both text embedding benchmarks (MTEB) and NLP benchmarks (MMLU). The results show that our proposed method significantly improves the original LLMs on MTEB while having a minimal impact on MMLU. Our strong results indicate that our approach can empower LLMs with state-of-the-art text embedding capabilities while maintaining their original NLP performance.
\end{abstract}

\section{Introduction}
\label{submission}
Large Language Models (LLMs) have revolutionized the field of Natural Language Processing (NLP), achieving remarkable success in various tasks such as text generation, question answering, language translation, and text summarization. However, their application in text embedding generation has been relatively slow due to the decoder-only attention mechanism employed by most LLMs. This limitation has led to the use of standalone embedding models, which can potentially hurt the performance of systems (e.g., retrieval augmented generation \cite{lewis_retrieval-augmented_2021}) that rely on both text embedding and generation capabilities.

Text embedding models play a crucial role in compressing long text sequences into semantic representations, facilitating downstream tasks like information retrieval, clustering, and classification. Those models, e.g., Bert \cite{devlin-etal-2019-bert}, OpenAI Embeddings \cite{neelakantan2022text}, RepLLaMA-RankLLaMA \cite{ma2024fine}, E5 \cite{wang2022text}, SimCSE \cite{gao_simcse_2022}, BGE \cite{bge_embedding}, mostly employ bidirectional attention to encode input text and extract embeddings using mean pooling or end-of-sentence (EOS) tokens. Those models are trained with masked language modeling and optionally further finetuned with contrastive learning. However, while effective for text embedding tasks, those models are unable to generate text directly. 

Recently Echo \cite{springer_repetition_2024} claimed decoder-only LLM could generate high quality embedding by appending a replication of input text after itself, at the cost of $4\times$ computation. GritLM \cite{muennighoff2024generative} perform fine-tuning with multi-task of self-supervised next token prediction with causal attention and supervised contrastive learning with bidirectional attention. The resultant model could be used for both embedding generation and language generation. Llm2vec \cite{behnamghader2024Llm2vec} finetunes a decoder-only LLM into an encoder using bidirectional attention and contrastive loss. Though Llm2vec reported SoTA performance on text embedding benchmarks MTEB, it completely lost text generation and reasoning capabilities.

To preserve the generation ability of LLMs, some other works \cite{muennighoff2022mteb} utilize encoder-decoder architecture, which uses encoder to encode input text into embedding and relies on decoder to generate the text from the encoded embedding (via cross attention). However, this encoder-decoder architecture is still lagging behind decoder-only model in NLP tasks \cite{warner_smarter_2024}.

Motivated by Gist \cite{mu_learning_2024} and VoCo-Llama \cite{ye2024voco}, our proposed method appends special token(s) to the input and train the model to compress the input into this special token. For training objectives, we continue to use next-token prediction. However, by controlling the attention mask, we ensure that the generation of text following the special token(s) cannot attend to the text preceding the special token(s). This approach encourages the model to compress information into the special token(s). For semantic embedding generation tasks, we can append the special token(s) to the end of the input text and extract the output from the last layer at the positions of these special token(s) to serve as the embedding. For text generation or reasoning tasks, the special token(s) are optional, allowing the model to function as a standard decoder-only language model. To further enhance embedding performance, we incorporate self-supervised contrastive learning as an additional training objective.

We evaluated the proposed method on models from Llama and Mistral families, ranging from 1B to 7B. The MTEB results indicates the proposed method dramatically enhanced the text embedding performances compared with original models and comparable to SoTA embedding models; while the proposed method only slightly degraded the performance on MMLU (language understanding). In addition, this is achieved by using as few as 32,000 rows of data.

In comparison to existing works, the advantages of the proposed method are as follows:
\begin{itemize}
\item Our method enables LLMs to generate high-quality embeddings of the input while preserving their original generation and reasoning capabilities. In contrast, Llm2vec \cite{behnamghader2024Llm2vec} converts the LLM into an encoder, resulting in the loss of language generation capability and limiting its functionality to a text embedding model.
\item Seamlessly compatible with all existing LLMs: our method can be integrated into post-training or finetuning stages of all existing LLMs.
\item Our method is computationally efficient. eliminating the need for additional parameters and massive training datasets. Notably, fine-tuning can converge with as few as 32,000 rows of training data. In  contrast, GritLM \cite{muennighoff2024generative} requires more than an order of magnitude ($10$ times) more data  for fine-tuning.
\item Our model is trained using a self-supervised approach, eliminating the need for extensive and costly labeling of data
\end{itemize}

% more recent work, noteLLM, VOCO-Llama, In-context Autoencoder for Context

% \begin{center}
% \textbf{\texttt{http://icml.cc/}}
% \end{center}

\section{Related Works}
\subsection{Large Language Models}
In recent years, large language models (LLM) have made tremendous progress, revolutionizing the field of natural language processing (NLP). These models have demonstrated remarkable capabilities in understanding and generating human-like text, achieving state-of-the-art results in various NLP tasks.

The most advanced large language models, e.g., GPT \cite{yenduri2023generativepretrainedtransformercomprehensive}, LLaMA 
 \cite{dubey2024llama}, Gemini \cite{team2023gemini}, Mistral \cite{jiang2023mistral}, Deepseek \cite{liu2024deepseek}, Qwen \cite{yang2024qwen2} are all based on decoder-only transformer. The decoder-only transformer architecture has proven to be a crucial factor in the success of these LLMs. This architecture uses causal attention, which allows the model to attend to previous tokens in the sequence while predicting the next token. The model is trained using a next token prediction loss function, which encourages the model to generate coherent and contextually relevant text.

These LLMs have demonstrated remarkable performance in instruction following, few-shot in-context learning, and reasoning. To mitigate knowledge cut-off, retrieval augmented generation (RAG) is developed, where the LLM could retrieve additional context from external knowledge base and then generate the response from combined query and retrieved context. In most recent work, researchers extend LLM and build agent, via tool utilization, external knowledge, enhanced reasoning and decision making capabilities as well as interaction with environment.
\subsection{Text Embedding Models}
Text embeddings are vector representations of text that encode its semantic information.
They are widely used in various natural language processing (NLP) tasks, such as information retrieval (IR), emotion detection, item recommendation, etc. Text embeddings are still widely used in retrieval and other tasks. For example, in retrieval augmented generation \cite{lewis2020retrieval}, context is first retrieved according to the embedding similarity to the query and then LLM generate the response based on both query and retrieved context. 

With decoder-only LLMs outperforming bidirectional encoders across a large variety of language understanding tasks, their impact on text embedding learning becomes more and more significant. Previous text embedding models are instead based on encoders using bidirectional attention. The encoders are usually trained with masked language modeling and optionally further finetuned with contrastive loss such as OpenAI Embeddings \cite{neelakantan2022text}, RepLLaMA-RankLLaMA \cite{ma2024fine}, E5 \cite{wang2022text}, SimCSE \cite{gao_simcse_2022}, BGE \cite{bge_embedding}. The embedding are usually extracted from the last hidden layer's output via mean pooling or (end of sentence) EOS token or other dedicated token(s). However those encoders model are found failing to capture the nuance differences in the input, e.g., it may generate similar embedding for "...I love apple..." vs "...I don't like apple...".

Some recent works tried to leverage the exceptional language capabilities of the pre-trained decoder-only LLMs by tuning them into text embedding models. Our method belongs to this line of work. Echo \cite{springer_repetition_2024} claimed decoder-only LLM could generate high quality embedding by appending a replication of input text after itself. However, this doubles the input length and thus results in $4\times$ computational costs. GritLM \cite{muennighoff2024generative} perform fine-tuning with multi-task of self-supervised next token prediction with causal attention and supervised contrastive learning with bidirectional attention. The resultant model could be used for both embedding generation and language generation. Llm2vec \cite{behnamghader2024Llm2vec} finetunes a decoder-only LLM into a strong text embedding model by a) converting causal attention to bidirection attention, b) masked token prediction and c) self-supervised contrastive learning. Though Llm2vec reported SoTA performance on text embedding benchmarks MTEB, it completely lost text generation and reasoning capabilities.
\subsection{Context Compression} 
Our work is also related to context compression for LLMs. The transformer used by LLMs is known to have quadratic computational complexity to the input token length due to the attention operation. However, the input is getting longer and longer as the LLMs become more capable, e.g., RAG would add retrieved context, chain of thought (CoT) needs to performance iterative thinking, more details and specifications are added to system prompt or instruction. Some works reduced the cost by making the transformer (attention) itself more efficient, e.g., Linformer \cite{wang2020linformerselfattentionlinearcomplexity}, Ring Attention \cite{liu2023ringattentionblockwisetransformers} and Attention Sink \cite{xiao2024efficientstreaminglanguagemodels}. 

The other works tried to reduce the input length by compressing the input or part of input (e.g., prompt) into fewer tokens. Gist \cite{mu_learning_2024} proposed to add gist tokens and fine tune the LLMs to compress the prompts to much shorter gist tokens (e.g., 4). ICAE \cite{ge_-context_2024} fine tuned an encoder to compress text to a few learnable tokens and use a frozen decoder (a pretrained LLM) to recover the text from the compressed tokens. SepLLM \cite{chen_sepllm_2024} and AutoCompressor \cite{chevalier_adapting_2023} took a similar method but extends to handle much longer input by first chunking the input into a few shorter segments and compress each segments sequentially. \cite{deng_silver_2024} studied how those compression methods work in different tasks of MTEB. It was found that those methods work notably well in tasks like RAG and long-document QA; but not yet reliable in rerank and synthetic recall task.
%%%%%%%%%%
\section{Proposed Method}

\begin{figure}[ht]
\vskip 0.2in
\begin{center}
\centerline{\includegraphics[width=\columnwidth]{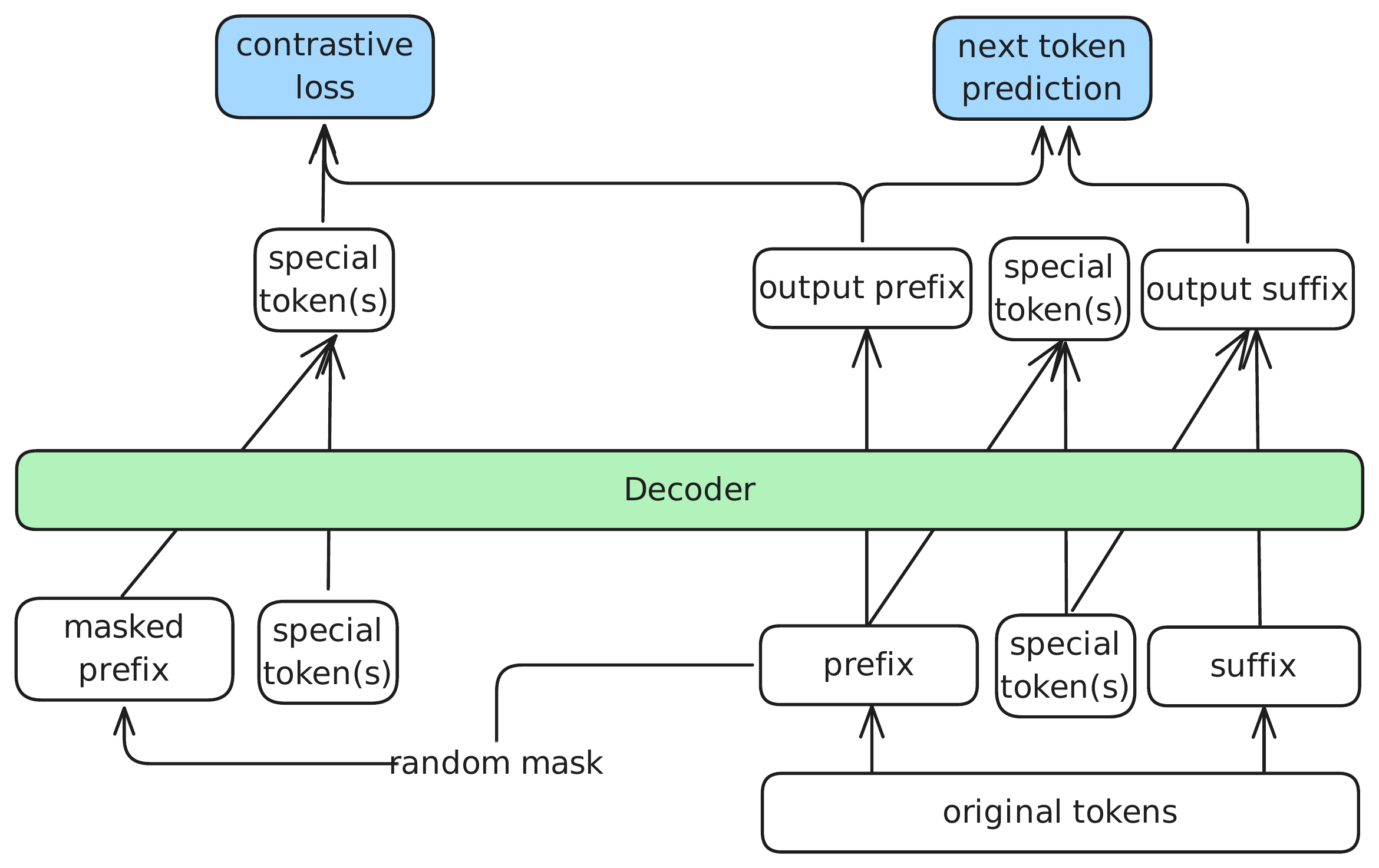}}
\caption{An overview of the proposed method. The proposed method takes tokenized text as input, then insert special tokens and performance next token prediction. In addition, it integrates contrastive loss by using embedding from random masked input as positive pairs.}
\label{llm_and_vec_arch}
\end{center}
\vskip -0.2in
\end{figure}

An overview of the proposed method is shown in Figure \ref{llm_and_vec_arch}. Given tokenized input text, we find a random position to insert the special token(s), which separate the input into prefix and suffix. The combined, prefix, special token(s) and suffix are fed into the decoder to perform next token prediction. To further enhance the embedding quality of the special token(s), a contrastive loss is applied by using the randomly masked prefix as positive pairs.

\subsection{Next Token Prediction (NTP) with Special Tokens}

In contrast to traditional NTP in LLMs, we use a hyperparameter, $p$, which governs the probability of inserting a special token into the input text. This approach enables us to perform a wide range of tasks, including:

\begin{itemize}
    \item Regular next token prediction: No special token(s) would be inserted and the model acts like other standard decoders.
    
    \item Text reconstruction \cite{ge2023context, springer_repetition_2024}:  In this task, the special token(s) are inserted after the entire input text, and the input text is then replicated following the special tokens. The model is thus required to compress the input text into the special token(s) and subsequently reconstruct it back to its original form.
    
    \item Text summary \cite{chevalier_adapting_2023}: The input consists of two parts: the context and a summary of that context. The special token(s) are strategically inserted between these two components, effectively separating the original text from its condensed representation. In this setup, the model is tasked with summarizing the input text into the special token(s), which are then decoded back into a textual form.
    
    \item Soft prompt \cite{mu_learning_2024}: The input comprises two components: the prompt and its corresponding answer. The special token(s) are inserted between these two elements. In this configuration, the model is tasked with representing the prompt as a concise set of special token(s), which then serve as the basis for generating a relevant and accurate response.
    
\end{itemize}

In our experiments, we consider \textit{regular next token prediction} and \textit{text reconstruction} due to the vast availability of the datasets. We will study the feasibility of other approaches in the future study.

With inputs in the form of `prefix special suffix', a constraint is imposed on the suffix tokens, explicitly preventing them from attending to the prefix tokens, similar to previous works \cite{ye2024voco}\cite{mu_learning_2024}. Consequently, the suffix tokens are only able to attend to themselves and the special token(s), effectively creating a "bottleneck" design. When multiple special tokens present, we further enforce each token is not able to see each other. This deliberate restriction ensures that the suffix tokens solely rely on the information encoded in the special tokens, rather than directly interacting with the prefix tokens, thereby achieving compression from the prefix tokens to the special tokens.

\begin{figure}[ht]
\vskip 0.2in
\begin{center}
\centerline{\includegraphics[width=\columnwidth]{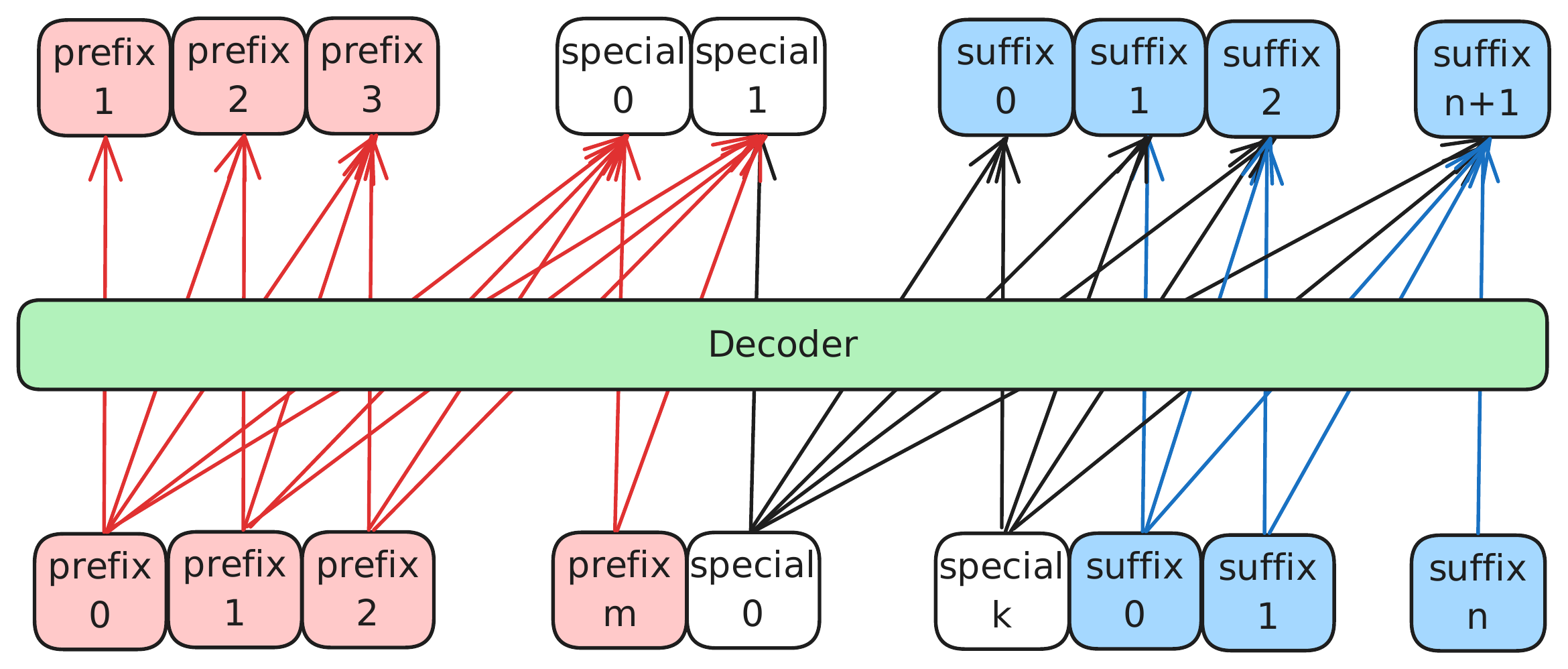}}
\caption{Illustration of the next token prediction with special tokens. The attention mask allows the suffix tokens after the special token attends to special and suffix tokens themselves but not any prefix tokens. This encourages the information of prefix tokens compressed into special tokens.}
\label{llm_and_vec_ntp}
\end{center}
\vskip -0.2in
\end{figure}

\begin{figure}[ht]
\vskip 0.2in
\begin{center}
\centerline{\includegraphics[width=\columnwidth]{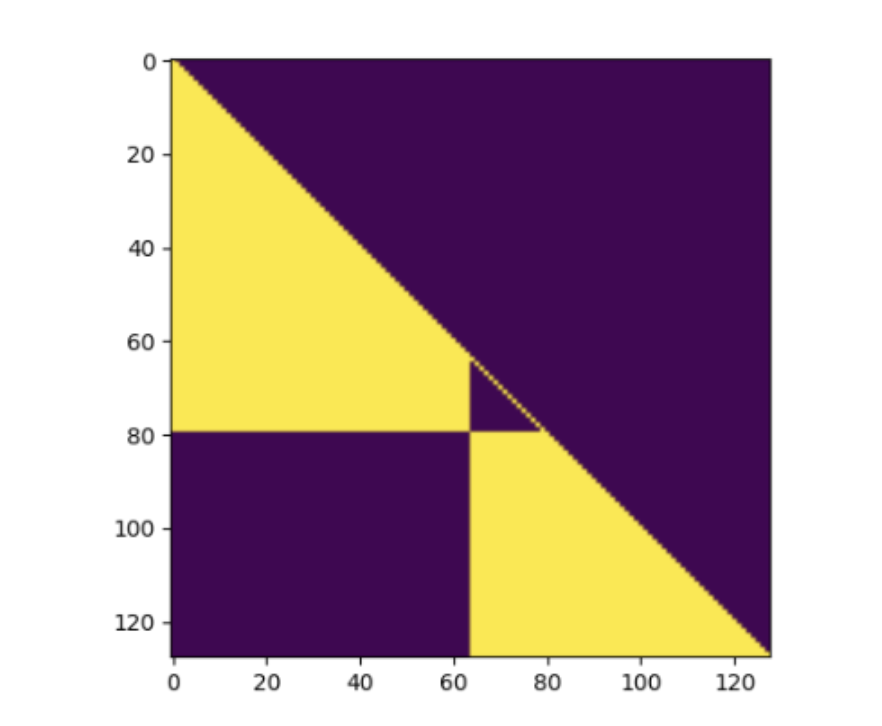}}
\caption{Illustration of the attention mask for adding k special tokens after input of m tokens. Here we have m tokens in prefix, k special tokens and n tokens in suffix. Yellow indicates row i could attend to column j; or blue otherwise. Note those special tokens don't attend to each other.}
\label{llm_and_vec_attention_mask}
\end{center}
\vskip -0.2in
\end{figure}

This design can be achieved by modifying the attention mask of a transformer. The attention mask is initialized as a lower left triangle, similar to casual mask used by a regular decoder-only transformer model. Next, we identify the bottleneck position of the special tokens, if present, and set mask off all the tokens preceding the first special token. This modification effectively creates a "bottleneck" in the attention mechanism, allowing the suffix tokens to only attend to themselves and the special tokens. We also disallow special tokens to attend to each other as we find this improves the performance.
A simple example illustrating the addition of k special tokens after an input of m tokens is shown in Figure \ref{llm_and_vec_attention_mask}. This figure demonstrates how the modified attention mask enables the desired bottleneck behavior.

As demonstrated in Section \ref{ablate_contrast}, incorporating our proposed NTP with special tokens yields a substantial improvement in the model's performance on MTEB (text embedding).

\subsubsection{Mixed NTP Loss}

\label{section:mixed_ntp}

Different from previous works \cite{ye2024voco, mu_learning_2024}, we propose to mix regular text (without inserting any special tokens) with `prefix special suffix` data in each batch with probability controlled by hyperparameter $p$. This mix mitigates catastrophic forgetting \cite{ibrahim_simple_2024} thus preserving original language capabilities of the decoder. In Section \ref{ablate_mix_ratio}, we performed an ablation study to evaluate the impact of this strategy and the hyperparamter $p$.

Specifically, for texts without special token, we just perform normal NTP as normal decoder-only LLM, which is useful to avoid catastrophic forgetting on its original LLM tasks. Since the insert of the special tokens are by random, we don't expect our model to fit where the special token resides in a string of texts. For texts with special tokens, we applies a loss mask to hold out special tokens and perform NTP on all other tokens.

\subsection{Contrastive Learning}\label{sec:contrast}
To further enhance the quality of embeddings for the special tokens, we also leveraged the contrastive loss. Contrastive learning is a popular paradigm to learn high quality text embeddings, e.g., DPR \cite{karpukhin_dense_2020}, SimCSE \cite{gao_simcse_2022}, CLIP \cite{radford_learning_2021}, Llm2vec \cite{behnamghader2024Llm2vec}. The contrastive loss could be performed in supervised way, if sets of positive pairs are available, otherwise self-supervised settings could be used.

In self-supervised setting, for each `prefix special suffix`, we perform a random dropout in prefix. `prefix special suffix` and its corresponding `dropout prefix special suffix` then form a positive pair; `prefix special suffix` and other `prefix special suffix` from the batch/dataset form the negative pairs. The model with contrastive loss is trained to maximize the similarity of the embedding of the special tokens from the positive pairs (e.g., `prefix special suffix` and its corresponding `dropout prefix special suffix`) and minimize the similarity of the embeddings from the negative pairs.
\begin{equation}
    \mathcal{L}_{cl}=\frac{e^{\lambda s(q,d^+)}}{e^{\lambda s(q,d^+)} + \sum_{d^-\in N}{e^{\lambda s(q,d^+)}}}
\end{equation}
where $s$ is similarity metric (e.g., cosine similarity) and $\lambda$ is learnable temperature parameter, $d^+$ and $d^-$ are the positive and negative pair to $q$. Following \cite{radford_learning_2021}, $\lambda$ is initialized as $\log{20}$ and clamped in $[0, \log{100}]$.

Different from \cite{behnamghader2024Llm2vec, muennighoff2024generative}, our model uses causal attention instead of bidirectional attention. This is critical to preserve the language capabilities of the model.

This contrastive loss is then combined with the proposed mixed NTP loss to train the model which results in the following final loss equation:
\begin{equation}
    \mathcal{L} = (1-\alpha)\mathcal{L}_{ntp} + \alpha \mathcal{L}_{cl}
\end{equation}
where $\alpha\in[0,1]$ controls the weight between those two losses. In our experiments, we found that starting with a small value of $\alpha$ (i.e., a larger weight on the next token prediction loss) and gradually increasing it in later stages (i.e., a larger weight on the contrastive loss) yields better results. Initializing with $\alpha=0$ (i.e., using only the mixed NTP loss) helps to prevent catastrophic forgetting, thereby preserving language understanding capabilities. Subsequently, we incrementally increase the weight on the contrastive loss to further enhance the embedding performance. In Section \ref{ablate_contrast}, we included the ablation study to evaluate the impact of adding contrastive loss.

If multiple special tokens are used, we could use mean pool or concatenation to aggregate the embeddings from those special tokens as final embedding. 

\section{Experiments}
In this section, we present results on both MTEB (text embedding benchmarks) and MMLU (language understanding). The proposed method is evaluated on two model families with model size from 1B to 8B. The result is compared with SoTA models.
\subsection{Tasks}
The Massive Text Embedding Benchmark (MTEB) \cite{muennighoff2022mteb} is a comprehensive benchmark designed to evaluate the quality of text embeddings across a wide range of tasks. It covers various tasks such as classification, clustering, ranking, and retrieval, and includes datasets in multiple languages. MTEB aims to provide a standardized way to assess the performance of text embedding models, making it easier to compare different models and approaches. We use the protocol and main score as metric (header of Table \ref{tab_full} according to \href{https://pypi.org/project/mteb/}{MTEB}).

MMLU (Multi-Task Multi-Lingual Understanding) \cite{hendrycks2021ethics,hendryckstest2021} is a benchmark designed to evaluate the performance of language models on a wide range of natural language understanding tasks across multiple languages. It's a comprehensive evaluation framework that assesses a model's ability to understand and process text in various contexts, including question answering, named entity recognition, part-of-speech tagging and so on. MMLU covers 57 languages and includes over 300 datasets, making it one of the most comprehensive benchmarks for evaluating language models. For MMLU, we follow \href{https://github.com/meta-llama/llama-models/blob/main/models/llama3_1/MODEL_CARD.md}{Llama 3.1 model card} and use five-shot and report macro\_avg/acc\_char.

\subsection{Set Up}
We use public dataset SIMCSE \cite{gao_simcse_2022} for our multi-task finetuning. No data of METB and MMLU would be exposed to our models during the finetuning. All the experiments are conducted in an self-supervised manner.

We use a batch size 32 and constraint the max sequence length to 512 for memory efficiency. In each training batch, we only insert the special token to $1-p=20\%$ of the raw text in each batch. We use learning rate $1e^{-4}$ for NTP fintuning and $1e^{-5}$ for contrastive learning with cosine scheduler. We apply start with $\alpha=0$ (NTP only) then apply increase to $\alpha=1$ (contrastive learning) after the first 100 iterations. We found as few as 32,000 rows of training data are needed, which makes our finetuning very efficient. All the hyper-parameters are obtained through the ablation studies on Llama 3-2-1B model and are applied across other LLM families. We finetune from the pretrain checked point in our experiments.
\subsection{Main results across various LLM families}
 We report the full results of applying our method to two LLM families Llama \cite{dubey2024llama} and Mistral \cite{jiang2023mistral} with model size from 1B to 8B in Table \ref{tab_full}. For comparison, we also report the result from existing SoTA models, especially echo \cite{springer_repetition_2024}, Llm2vec \cite{behnamghader2024Llm2vec} and GritLM \cite{muennighoff2024generative}, which are tuned from decoder-only LLM.

The result is reported on both MTEB (embedding benchmark) and MMLU (language understanding benchmark). For MTEB we follow the protocol of \cite{behnamghader2024Llm2vec} for metrics and instructions. For MMLU, we follow \href{https://github.com/meta-llama/llama-models/blob/main/models/llama3_1/MODEL_CARD.md}{Llama 3.1 model card} and use five-shot and report macro\_avg/acc\_char. To extract embedding from vanilla Llama or Mistral model, we apply mean pooling on the last layer's output, which was also used in \cite{behnamghader2024Llm2vec}.

In our experiment, we were not able to fully reproduce the MTEB results for two baseline models Llama 3 8B and Mistral instruct v0.2 7B from \cite{behnamghader2024Llm2vec, muennighoff2024generative}, possibly due to differences in the protocol or changes of access to the benchmark (for which we are still working on). Thus we reported both our reproduced results and cited the results from \cite{behnamghader2024Llm2vec, muennighoff2024generative} for those two models for reference.

Table \ref{tab_full} indicates after applying the proposed method on Llama and Mistral models, their performance on MTEB got significantly improved, e.g., the retrieval task is improved from $4.38$ to $25.6$ for Llama 3.2 1B. In the meanwhile the performance on MMLU is only slightly degraded. This has shown the effectiveness of the proposed method across different model families and model size.

When comparing to other SoTA methods, our method achieves comparable results with Llm2vec on Llama 1B and Mistral 7B on embedding tasks; while lagging behind on Llama 8B. However, with Llm2vec, the model lost the language generation capabilities. In comparing with GritLM, our method also achieves a competitive result on both MTEB and MMLU, while only requires $<\frac{1}{10}$ of data for finetuning.

\begin{table*}[t]
\caption{The performance of proposed method on MTEB (text embedding benchmark) and MMLU (multilingual multitask language understanding benchmark), with comparison to Llm2vec and other LLMs. For MTEB, we reported average of each group of tasks. Pair and CLS is pair-classification and classification task of MTEB respectively. The scores are the higher the better. For MMLU, we followed \href{https://github.com/meta-llama/llama-models/blob/main/models/llama3_1/MODEL_CARD.md}{Llama 3.1 model card}, use five-shot and report macro\_avg/acc\_char. [1] The results reported from \cite{muennighoff2024generative}; [2]The results reported from \cite{behnamghader2024Llm2vec} Table 1. We couldn't fully reproduce the Llama 3 8B and Mistral Instruct v0.2 results on MTEB reported from those two papers and we reported our own results in this table as a reference.}
\label{tab_full}
\vskip 0.15in
\begin{center}
\begin{small}
\begin{sc}
\begin{tabular}{p{3.6cm}|p{1.2cm}p{1.2cm}p{1.2cm}p{1.2cm}p{1.2cm}p{1.2cm}p{1.2cm}|p{1.2cm}}
\toprule
models &           Cluster vMeas. & Rerank MAP & Pair APCosine & STS Spear   & Retr ndcg\@10 & SummEval Spear & CLS Acc & MMLU \\
\midrule
% BERT \todo{\# parameter?}              & 30.21   & 43.44   & 56.33               & 54.36 & 10.56  & 29.81  & 61.66 & N.A.  \\ \midrule
S-Llama-1.3B[2] & 28.02 & 38.02 & 42.19 & 49.15 & 9.47 & 24.98 & 59.79 & 25.71 \\
echo S-Llama[2] & 30.3 & 40.52 & 52.08 & 59.36 & 10.36 & 22.79&63.75 & N.A.\\
Llm2vec S-Llama[2] & 37.45 & 47.70 & 72.21 & 71.61& 25.93 & 31.23 & 67.67 & N.A. \\
\rowcolor[gray]{.9}
Llama 3-2-1B       & 23.11   & 33.90   & 29.89               & 33.32 & 4.38   & 23.5   & 40.26 & 31.7\\
\rowcolor[gray]{.9}
GEM Llama 3.1-1B & 40.59   & 53.34   & 64.88               & 68.42 & 30.10  &  25.6  & 58.78 & 28.36\\\midrule
\rowcolor[gray]{.9}
LLAMA 3-2-3B &25.31 & 39.67 & 35.57 & 33.28 & 8.16 & 20.36 & 35.99 &  58\\
\rowcolor[gray]{.9}
GEM LLAMA 3-2-3B &  42.68  & 63.41  & 78.14                & 74.91 & 38.14  & 24.73  & 63.19  & 54.3 \\ 
\midrule
% Llm2vec Llama 2-7B & 38.11   & 52.95   & 77.88               & 76.43 & 36.76  & 31.38  & 71.56 & N.A.\\
Llama 3-8B[2] & 36.84& 46.22& 60.94 & 62.80& 15.17 & 25.51& 67.41 & 66.7\\
Llm2vec Llama 3-8B[2] & 41.00   & 53.09   & 87.80                  & 83.58    & 39.19  & 30.94     & 71.88 & N.A.\\ 
\rowcolor[gray]{.9}
Llama 3-8B    & 25.57 & 42.80 & 42.53 & 40.50 & 11.86 & 24.37  & 41.36 & 66.7 \\
\rowcolor[gray]{.9}
GEM Llama 3-8B & 41.58   & 60.10  &  77.95 & 69.00 & 34.08  & 31.11 &   66.3 & 57.24 \\ \midrule
% please merge the best results of the two rows below into the row above
% GEM Llama 3-8B & 34.94   & 51.53  &  60.46               & 54.73 & 16.23  & 31.11 &  53.77 & 57.24 \\ 
% GEM Llama 3-8B ins & TBD & 60.10 & 77.95 & 69.00 & TBD & 29.36 & TBD & 57.24\\ \midrule
Mistral 7B [1,2] & 34.6 & 44.8 & 59.6 & 63.4& 16.3& 25.9 & 67.1 & 53.0\\
GritLM 7B[1] & 50.6 & 60.5 & 87.2 & 83.4 & 57.4 & 30.4 & 79.5 & 57.6\\
echo Mistral 7B[2] & 36.78 & 51.07 & 75.87 & 73.60 & 22.68 & 29.56&72.69 & N.A.\\
Llm2vec Mistral 7B[2] & 40.63 & 53.99 & 80.94 & 78.50 & 38.05 & 30.19 & 74.07  & N.A. \\
\rowcolor[gray]{.9}
Mistral 7B & 25.74 & 43.12 & 38.71 & 36.41& 10.64& 23.84 & 42.14 & 44.4\\
\rowcolor[gray]{.9}
GEM Mistral 7B & 32.53& 55.56& 67.59& 67.95& 26.53 & 29.36 & 56.92 & 36.2 \\
% normalized & 41.39 & 57.24 & 88.48 & 94.94 & 32.19 & 31.42 & 97.84 & \todo{TBD} \\
\bottomrule
\end{tabular}
\end{sc}
\end{small}
\end{center}
\vskip -0.1in
\end{table*}

\section{Ablation studies}
In this section, we performed ablation studies to evaluate the impact of different components of the proposed method. To save computation, we focus on Llama 3.2 1B model and evaluated on MMLU and $15$ sub-tasks of MTEB, following \cite{behnamghader2024Llm2vec}. The definition of $15$ subtasks can be found in Appendix \ref{appendix:15_tasks}. 

\subsection{The mix ratio of special token}\label{ablate_mix_ratio}
In Section \ref{section:mixed_ntp} we finetune the original model with a mix of raw texts with probability $p$ and text inserted with special tokens with probability $1-p$. In the experiment, it is found this achieves good embedding quality for the special token while mitigates the catastrophic forgetting of the original language capabilities. In this section, we performed an ablation study on the impact of this mix ratio $p$. The average results on key $15$ sub tasks of MTEB and the macro average of acc char of MMLU are shown in table \ref{tab_ratio}.

\begin{table}[t]
\caption{The average of $15$ MTEB subtasks' scores with respect to different mix ratios of raw text and text inserted with special tokens. The scores are the higher the better. We also present the macro average acc char of MMLU.}
\label{tab_ratio}
\vskip 0.15in
\begin{center}
\begin{small}
\begin{sc}
\begin{tabular}{lcc}
\toprule
mix rate $p$ & MTEB & MMLU \\
\midrule
0.0    & 41.24 & 26.08 \\
0.2 & 41.10  & 26.72 \\
0.4 &  37.74 & 27.01 \\
0.6 & 45.98  & 27.24\\ 
0.8 & 48.11  & 27.72 \\
0.9 & 44.71 & 27.5 \\
0.95 & 40.23 & 28.58 \\ 
0.99 & 27.74 & 29.77 \\
\bottomrule
\end{tabular}
\end{sc}
\end{small}
\end{center}
\vskip -0.1in
\end{table}

As indicated by the table, increasing the ratio of raw text helps reserve the model's performance on MMLU. This could be explained by replaying the data used in previous pretrain or finetuning could help mitigate the catastrophic forgetting of LLM \cite{ibrahim_simple_2024, guo2024efficientcontinualpretrainingmitigating}. However, after $p>0.8$, the performance in MTEB begins to degrade, i.e., the model struggled to adapt to new tasks. As $p$ got too large and the texts with special tokens became too scarce, then it is difficult for the model to learn how to "compress" the prefix text the special tokens. Based on those findings, we use $p=0.8$ in our experiments.

\subsection{Contrastive loss}\label{ablate_contrast}
We proposed multi-tasking of mixed NTP loss and contrastive loss in Section \ref{sec:contrast}. In this section, we described the ablation study to evaluate the impact of those two training objectives. In this study, we start with $\alpha=0$ (mixed NTP only), then quickly increase $\alpha \rightarrow 1$ (contrastive loss) after the first 100 iterations. 

The results are shown in table \ref{tab_contra}, which clearly shows our mixed NTP loss could significantly improve the model's performance on MTEB while have a small degradation on MMLU; adding contrastive loss could further improve the performance in MTEB as well as preserving the LLM's performance in MMLU benchmark. Thus we use multi-tasking of mixed NTP loss and contrastive loss as our proposed recipes.

\begin{table}[t]
\caption{Ablation study on the proposed training objectives : mixed NTP loss and contrastive loss. The performance is measured by the average of $15$ MTEB subtasks' scores as well as MMLU.}
\label{tab_contra}
\vskip 0.15in
\begin{center}
\begin{small}
\begin{sc}
\begin{tabular}{rcc}
\toprule
Training Objetives & MTEB & MMLU\\
\midrule
Vanilla LlaMA 3.2 1B     & 18.29 & 31.7 \\
+Mixed NTP    & 48.11 & 27.72  \\
+Mixed NTP+Contrastive & 54.35  & 28.36 \\
\bottomrule
\end{tabular}
\end{sc}
\end{small}
\end{center}
\vskip -0.1in
\end{table}

\subsection{Number of special tokens}
In this section, we study the influence of different number of special tokens. According to table \ref{tab_toks}, increasing number of special tokens (from $1$ to $10$) significantly increase model's performance on retrieval tasks (MTEB). However, the performance of MMLU is not affected too much with the choices of number of special tokens. This is expected as the special tokens are not used in MMLU after the model is trained.

\begin{table}[t]
\caption{Ablation study on the number of special tokens. The performance is measured by the average of $15$ MTEB subtasks' scores as well as MMLU.}
\label{tab_toks}
\vskip 0.15in
\begin{center}
\begin{small}
\begin{sc}
\begin{tabular}{rcc}
\toprule
Training Objetives & MTEB & MMLU\\
\midrule
1 token     & 54,35 & 28.36 \\
2 tokens    & 54.67 & 28.48  \\
5 tokens & 55.88  & 29.81 \\
10 tokens & 56.12  & 28.78 \\
\bottomrule
\end{tabular}
\end{sc}
\end{small}
\end{center}
\vskip -0.1in
\end{table}

% \todo{should we describe or study the impact of multi-task weight $\alpha$ as well? which essentially is the \# of data used for each step.}
\subsection{Scaling rule}\label{ablate_scale}
In this section, we apply the proposed method to Llama families at three different parameters scales e.g. Llama 3-2-1B, Llama 3-2-3B, Llama 3-1-8B and evaluate their performance. The results of MTEB and MMLU are shown in table \ref{tab_scale}.  

As shown in the table, the proposed method could be applied to Llama at different scales and the performance on MTEB and MMLU increases with the model's scale. However, on MTEB Llama 3-8B model didn't outperform 3B model significantly, which may suggest Llama 3-8B would need different hyper-parameters.  

As for MMLU, we find the smaller models preserve the original model's performance better. In our 1B and 3B finetuned model, they almost keep the same reasoning ability as the original one. However, the 8B finetuned model yields a significant drop. This could also suggest we would need different hyper-parameter for Llama 3-8B. This would be studied in future works.
\begin{table}[t]
\caption{Comparison of applying proposed method on Llama models at varying scales (1B, 3B and 8B) on MTEB and MMLU. The original Llama model's performance is also provided as baseline. The MTEB result is averaged over $15$ tasks. }
\label{tab_scale}
\vskip 0.15in
\begin{center}
\begin{small}
\begin{sc}
\begin{tabular}{r|cc|cc}
\toprule
Model & \multicolumn{2}{|c|}{MTEB} & \multicolumn{2}{|c}{MMLU} \\
Size & baseline & GEM & \href{https://github.com/meta-llama/llama-models/blob/main/models/llama3_2/MODEL_CARD.md}{baseline} & GEM \\
\midrule
Llama 3.2-1B & 18.29    & 54.35 & 31.7 &28.3 \\
Llama 3.2-3B  &  21.60 & 59.06 & 58 & 54.30 \\
LLama 3-8B & 26.67 & 54.48 & 66.7 & 57.24\\
\bottomrule
\end{tabular}
\end{sc}
\end{small}
\end{center}
\vskip -0.1in
\end{table}
% \subsection{Number of special tokens}
% \todo{Let us think what could be put here.} In MTEB tasks, we didn't see significant gains by increase the number of tokens in our prompt.
\section{Conclusion}
In this paper, we proposed a method that significantly improves the text embedding performance of an LLM while preserving the language generation capabilities. Our method is very efficient - just utilizing 32,000 rows of training data. This method is evaluated on models from two families (Lllam and Mistral) with model scaling from 1B to 8B, demonstrating its general applicability to different LLMs. We believe this method could be integrated into post-training and finetuning stage of existing LLMs. However, we found the proposed method is less effective on Llama 3 8B models, which could be suboptimal hyper-parameter. We plan to address it in our future work.
\clearpage
\section{Impact Statement}
This paper presents a novel approach to enhancing the text embedding capabilities of large language models (LLMs) while preserving their generation and reasoning abilities. The proposed method has the potential to significantly impact various applications of LLMs, including natural language processing, information retrieval, and conversational AI. The broader societal implications of this work are multifaceted:

\begin{itemize}
    \item \textbf{Improved accessibility}: By enabling LLMs to generate high-quality embeddings, we can improve the accessibility of complex information for individuals with varying levels of literacy and cognitive abilities.
    \item \textbf{Enhanced decision-making}: The ability of LLMs to reason and make decisions based on complex conditions and scenarios can lead to more informed decision-making in various domains, such as healthcare, finance, and education.
    \item \textbf{Increased efficiency}: The proposed method can streamline the process of generating text embeddings, reducing the computational resources required and making it more feasible to deploy LLMs in real-world applications.
\end{itemize}

However, there are also potential risks and challenges associated with this technology:
\begin{itemize}
    \item \textbf{Bias and fairness}: As with any machine learning model, there is a risk of bias and unfairness in the generated embeddings, which could perpetuate existing social inequalities.
    \item \textbf{Misuse and manipulation}: The ability of LLMs to generate convincing text could be misused for malicious purposes, such as spreading misinformation or propaganda.
\end{itemize}

To mitigate these risks, we emphasize the importance of responsible development and deployment of this technology, including:
\begin{itemize}
    \item \textbf{Regular auditing and testing}: Regularly auditing and testing the proposed method to ensure that it is fair, transparent, and unbiased.
    \item \textbf{Human oversight and review}: Implementing human oversight and review processes to detect and correct any potential biases or errors in the generated embeddings.
    \item \textbf{Education and awareness}: Educating users and developers about the potential risks and benefits of this technology and promoting responsible use and deployment.
\end{itemize}

By acknowledging these potential implications and taking steps to address them, we hope to contribute to the development of more responsible and beneficial AI technologies.
% \vfill
\bibliography{template.bib}

\begin{thebibliography}{35}
\providecommand{\natexlab}[1]{#1}
\providecommand{\url}[1]{\texttt{#1}}
\expandafter\ifx\csname urlstyle\endcsname\relax
  \providecommand{\doi}[1]{doi: #1}\else
  \providecommand{\doi}{doi: \begingroup \urlstyle{rm}\Url}\fi

\bibitem[Lewis et~al.(2021)Lewis, Perez, Piktus, Petroni, Karpukhin, Goyal,
  Küttler, Lewis, Yih, Rocktäschel, Riedel, and
  Kiela]{lewis_retrieval-augmented_2021}
Patrick Lewis, Ethan Perez, Aleksandra Piktus, Fabio Petroni, Vladimir
  Karpukhin, Naman Goyal, Heinrich Küttler, Mike Lewis, Wen-tau Yih, Tim
  Rocktäschel, Sebastian Riedel, and Douwe Kiela.
\newblock Retrieval-{Augmented} {Generation} for {Knowledge}-{Intensive} {NLP}
  {Tasks}, April 2021.
\newblock URL \url{http://arxiv.org/abs/2005.11401}.
\newblock arXiv:2005.11401 [cs].

\bibitem[Devlin et~al.(2019)Devlin, Chang, Lee, and
  Toutanova]{devlin-etal-2019-bert}
Jacob Devlin, Ming-Wei Chang, Kenton Lee, and Kristina Toutanova.
\newblock {BERT}: Pre-training of deep bidirectional transformers for language
  understanding.
\newblock In Jill Burstein, Christy Doran, and Thamar Solorio, editors,
  \emph{Proceedings of the 2019 Conference of the North {A}merican Chapter of
  the Association for Computational Linguistics: Human Language Technologies,
  Volume 1 (Long and Short Papers)}, pages 4171--4186, Minneapolis, Minnesota,
  June 2019. Association for Computational Linguistics.
\newblock \doi{10.18653/v1/N19-1423}.
\newblock URL \url{https://aclanthology.org/N19-1423}.

\bibitem[Neelakantan et~al.(2022)Neelakantan, Xu, Puri, Radford, Han, Tworek,
  Yuan, Tezak, Kim, Hallacy, et~al.]{neelakantan2022text}
Arvind Neelakantan, Tao Xu, Raul Puri, Alec Radford, Jesse~Michael Han, Jerry
  Tworek, Qiming Yuan, Nikolas Tezak, Jong~Wook Kim, Chris Hallacy, et~al.
\newblock Text and code embeddings by contrastive pre-training.
\newblock \emph{arXiv preprint arXiv:2201.10005}, 2022.

\bibitem[Ma et~al.(2024)Ma, Wang, Yang, Wei, and Lin]{ma2024fine}
Xueguang Ma, Liang Wang, Nan Yang, Furu Wei, and Jimmy Lin.
\newblock Fine-tuning llama for multi-stage text retrieval.
\newblock In \emph{Proceedings of the 47th International ACM SIGIR Conference
  on Research and Development in Information Retrieval}, pages 2421--2425,
  2024.

\bibitem[Wang et~al.(2022)Wang, Yang, Huang, Jiao, Yang, Jiang, Majumder, and
  Wei]{wang2022text}
Liang Wang, Nan Yang, Xiaolong Huang, Binxing Jiao, Linjun Yang, Daxin Jiang,
  Rangan Majumder, and Furu Wei.
\newblock Text embeddings by weakly-supervised contrastive pre-training.
\newblock \emph{arXiv preprint arXiv:2212.03533}, 2022.

\bibitem[Gao et~al.(2022)Gao, Yao, and Chen]{gao_simcse_2022}
Tianyu Gao, Xingcheng Yao, and Danqi Chen.
\newblock {SimCSE}: {Simple} {Contrastive} {Learning} of {Sentence}
  {Embeddings}, May 2022.
\newblock URL \url{http://arxiv.org/abs/2104.08821}.
\newblock arXiv:2104.08821 [cs].

\bibitem[Xiao et~al.(2023)Xiao, Liu, Zhang, and Muennighoff]{bge_embedding}
Shitao Xiao, Zheng Liu, Peitian Zhang, and Niklas Muennighoff.
\newblock C-pack: Packaged resources to advance general chinese embedding,
  2023.

\bibitem[Springer et~al.(2024)Springer, Kotha, Fried, Neubig, and
  Raghunathan]{springer_repetition_2024}
Jacob~Mitchell Springer, Suhas Kotha, Daniel Fried, Graham Neubig, and Aditi
  Raghunathan.
\newblock Repetition {Improves} {Language} {Model} {Embeddings}, February 2024.
\newblock URL \url{http://arxiv.org/abs/2402.15449}.
\newblock arXiv:2402.15449 [cs].

\bibitem[Muennighoff et~al.(2024)Muennighoff, Su, Wang, Yang, Wei, Yu, Singh,
  and Kiela]{muennighoff2024generative}
Niklas Muennighoff, Hongjin Su, Liang Wang, Nan Yang, Furu Wei, Tao Yu,
  Amanpreet Singh, and Douwe Kiela.
\newblock Generative representational instruction tuning.
\newblock \emph{arXiv preprint arXiv:2402.09906}, 2024.

\bibitem[BehnamGhader et~al.(2024)BehnamGhader, Adlakha, Mosbach, Bahdanau,
  Chapados, and Reddy]{behnamghader2024Llm2vec}
Parishad BehnamGhader, Vaibhav Adlakha, Marius Mosbach, Dzmitry Bahdanau,
  Nicolas Chapados, and Siva Reddy.
\newblock Llm2vec: Large language models are secretly powerful text encoders.
\newblock \emph{arXiv preprint arXiv:2404.05961}, 2024.

\bibitem[Muennighoff et~al.(2022)Muennighoff, Tazi, Magne, and
  Reimers]{muennighoff2022mteb}
Niklas Muennighoff, Nouamane Tazi, Lo{\"\i}c Magne, and Nils Reimers.
\newblock {MTEB}: Massive text embedding benchmark.
\newblock \emph{arXiv preprint arXiv:2210.07316}, 2022.

\bibitem[Warner et~al.(2024)Warner, Chaffin, Clavié, Weller, Hallström,
  Taghadouini, Gallagher, Biswas, Ladhak, Aarsen, Cooper, Adams, Howard, and
  Poli]{warner_smarter_2024}
Benjamin Warner, Antoine Chaffin, Benjamin Clavié, Orion Weller, Oskar
  Hallström, Said Taghadouini, Alexis Gallagher, Raja Biswas, Faisal Ladhak,
  Tom Aarsen, Nathan Cooper, Griffin Adams, Jeremy Howard, and Iacopo Poli.
\newblock Smarter, {Better}, {Faster}, {Longer}: {A} {Modern} {Bidirectional}
  {Encoder} for {Fast}, {Memory} {Efficient}, and {Long} {Context} {Finetuning}
  and {Inference}, December 2024.
\newblock URL \url{http://arxiv.org/abs/2412.13663}.
\newblock arXiv:2412.13663 [cs].

\bibitem[Mu et~al.(2024)Mu, Li, and Goodman]{mu_learning_2024}
Jesse Mu, Xiang~Lisa Li, and Noah Goodman.
\newblock Learning to {Compress} {Prompts} with {Gist} {Tokens}, February 2024.
\newblock URL \url{http://arxiv.org/abs/2304.08467}.
\newblock arXiv:2304.08467 [cs].

\bibitem[Ye et~al.(2024)Ye, Gan, Huang, Ge, Shan, and Tang]{ye2024voco}
Xubing Ye, Yukang Gan, Xiaoke Huang, Yixiao Ge, Ying Shan, and Yansong Tang.
\newblock Voco-llama: Towards vision compression with large language models.
\newblock \emph{arXiv preprint arXiv:2406.12275}, 2024.

\bibitem[Yenduri et~al.(2023)Yenduri, M, G, Y, Srivastava, Maddikunta, G,
  Jhaveri, B, Wang, Vasilakos, and
  Gadekallu]{yenduri2023generativepretrainedtransformercomprehensive}
Gokul Yenduri, Ramalingam M, Chemmalar~Selvi G, Supriya Y, Gautam Srivastava,
  Praveen Kumar~Reddy Maddikunta, Deepti~Raj G, Rutvij~H Jhaveri, Prabadevi B,
  Weizheng Wang, Athanasios~V. Vasilakos, and Thippa~Reddy Gadekallu.
\newblock Generative pre-trained transformer: A comprehensive review on
  enabling technologies, potential applications, emerging challenges, and
  future directions, 2023.
\newblock URL \url{https://arxiv.org/abs/2305.10435}.

\bibitem[Dubey et~al.(2024)Dubey, Jauhri, Pandey, Kadian, Al-Dahle, Letman,
  Mathur, Schelten, Yang, Fan, et~al.]{dubey2024llama}
Abhimanyu Dubey, Abhinav Jauhri, Abhinav Pandey, Abhishek Kadian, Ahmad
  Al-Dahle, Aiesha Letman, Akhil Mathur, Alan Schelten, Amy Yang, Angela Fan,
  et~al.
\newblock The llama 3 herd of models.
\newblock \emph{arXiv preprint arXiv:2407.21783}, 2024.

\bibitem[Team et~al.(2023)Team, Anil, Borgeaud, Alayrac, Yu, Soricut,
  Schalkwyk, Dai, Hauth, Millican, et~al.]{team2023gemini}
Gemini Team, Rohan Anil, Sebastian Borgeaud, Jean-Baptiste Alayrac, Jiahui Yu,
  Radu Soricut, Johan Schalkwyk, Andrew~M Dai, Anja Hauth, Katie Millican,
  et~al.
\newblock Gemini: a family of highly capable multimodal models.
\newblock \emph{arXiv preprint arXiv:2312.11805}, 2023.

\bibitem[Jiang et~al.(2023)Jiang, Sablayrolles, Mensch, Bamford, Chaplot,
  Casas, Bressand, Lengyel, Lample, Saulnier, et~al.]{jiang2023mistral}
Albert~Q Jiang, Alexandre Sablayrolles, Arthur Mensch, Chris Bamford,
  Devendra~Singh Chaplot, Diego de~las Casas, Florian Bressand, Gianna Lengyel,
  Guillaume Lample, Lucile Saulnier, et~al.
\newblock Mistral 7b.
\newblock \emph{arXiv preprint arXiv:2310.06825}, 2023.

\bibitem[Liu et~al.(2024)Liu, Feng, Xue, Wang, Wu, Lu, Zhao, Deng, Zhang, Ruan,
  et~al.]{liu2024deepseek}
Aixin Liu, Bei Feng, Bing Xue, Bingxuan Wang, Bochao Wu, Chengda Lu, Chenggang
  Zhao, Chengqi Deng, Chenyu Zhang, Chong Ruan, et~al.
\newblock Deepseek-v3 technical report.
\newblock \emph{arXiv preprint arXiv:2412.19437}, 2024.

\bibitem[Yang et~al.(2024)Yang, Yang, Zhang, Hui, Zheng, Yu, Li, Liu, Huang,
  Wei, et~al.]{yang2024qwen2}
An~Yang, Baosong Yang, Beichen Zhang, Binyuan Hui, Bo~Zheng, Bowen Yu,
  Chengyuan Li, Dayiheng Liu, Fei Huang, Haoran Wei, et~al.
\newblock Qwen2. 5 technical report.
\newblock \emph{arXiv preprint arXiv:2412.15115}, 2024.

\bibitem[Lewis et~al.(2020)Lewis, Perez, Piktus, Petroni, Karpukhin, Goyal,
  K{\"u}ttler, Lewis, Yih, Rockt{\"a}schel, et~al.]{lewis2020retrieval}
Patrick Lewis, Ethan Perez, Aleksandra Piktus, Fabio Petroni, Vladimir
  Karpukhin, Naman Goyal, Heinrich K{\"u}ttler, Mike Lewis, Wen-tau Yih, Tim
  Rockt{\"a}schel, et~al.
\newblock Retrieval-augmented generation for knowledge-intensive nlp tasks.
\newblock \emph{Advances in Neural Information Processing Systems},
  33:\penalty0 9459--9474, 2020.

\bibitem[Wang et~al.(2020)Wang, Li, Khabsa, Fang, and
  Ma]{wang2020linformerselfattentionlinearcomplexity}
Sinong Wang, Belinda~Z. Li, Madian Khabsa, Han Fang, and Hao Ma.
\newblock Linformer: Self-attention with linear complexity, 2020.
\newblock URL \url{https://arxiv.org/abs/2006.04768}.

\bibitem[Liu et~al.(2023)Liu, Zaharia, and
  Abbeel]{liu2023ringattentionblockwisetransformers}
Hao Liu, Matei Zaharia, and Pieter Abbeel.
\newblock Ring attention with blockwise transformers for near-infinite context,
  2023.
\newblock URL \url{https://arxiv.org/abs/2310.01889}.

\bibitem[Xiao et~al.(2024)Xiao, Tian, Chen, Han, and
  Lewis]{xiao2024efficientstreaminglanguagemodels}
Guangxuan Xiao, Yuandong Tian, Beidi Chen, Song Han, and Mike Lewis.
\newblock Efficient streaming language models with attention sinks, 2024.
\newblock URL \url{https://arxiv.org/abs/2309.17453}.

\bibitem[Ge et~al.(2024)Ge, Hu, Wang, Wang, Chen, and Wei]{ge_-context_2024}
Tao Ge, Jing Hu, Lei Wang, Xun Wang, Si-Qing Chen, and Furu Wei.
\newblock In-context {Autoencoder} for {Context} {Compression} in a {Large}
  {Language} {Model}, May 2024.
\newblock URL \url{http://arxiv.org/abs/2307.06945}.
\newblock arXiv:2307.06945 [cs].

\bibitem[Chen et~al.(2024)Chen, Shi, Li, Gao, Ren, Chen, Jiang, Li, Liu, and
  Huang]{chen_sepllm_2024}
Guoxuan Chen, Han Shi, Jiawei Li, Yihang Gao, Xiaozhe Ren, Yimeng Chen, Xin
  Jiang, Zhenguo Li, Weiyang Liu, and Chao Huang.
\newblock {SepLLM}: {Accelerate} {Large} {Language} {Models} by {Compressing}
  {One} {Segment} into {One} {Separator}, December 2024.
\newblock URL \url{http://arxiv.org/abs/2412.12094}.
\newblock arXiv:2412.12094 [cs].

\bibitem[Chevalier et~al.(2023)Chevalier, Wettig, Ajith, and
  Chen]{chevalier_adapting_2023}
Alexis Chevalier, Alexander Wettig, Anirudh Ajith, and Danqi Chen.
\newblock Adapting {Language} {Models} to {Compress} {Contexts}, November 2023.
\newblock URL \url{http://arxiv.org/abs/2305.14788}.
\newblock arXiv:2305.14788 [cs].

\bibitem[Deng et~al.(2024)Deng, Zhang, Mao, Li, Huang, Yu, and
  Dou]{deng_silver_2024}
Chenlong Deng, Zhisong Zhang, Kelong Mao, Shuaiyi Li, Xinting Huang, Dong Yu,
  and Zhicheng Dou.
\newblock A {Silver} {Bullet} or a {Compromise} for {Full} {Attention}? {A}
  {Comprehensive} {Study} of {Gist} {Token}-based {Context} {Compression},
  December 2024.
\newblock URL \url{http://arxiv.org/abs/2412.17483}.
\newblock arXiv:2412.17483 [cs].

\bibitem[Ge et~al.(2023)Ge, Hu, Wang, Wang, Chen, and Wei]{ge2023context}
Tao Ge, Jing Hu, Lei Wang, Xun Wang, Si-Qing Chen, and Furu Wei.
\newblock In-context autoencoder for context compression in a large language
  model.
\newblock \emph{arXiv preprint arXiv:2307.06945}, 2023.

\bibitem[Ibrahim et~al.(2024)Ibrahim, Thérien, Gupta, Richter, Anthony,
  Lesort, Belilovsky, and Rish]{ibrahim_simple_2024}
Adam Ibrahim, Benjamin Thérien, Kshitij Gupta, Mats~L. Richter, Quentin
  Anthony, Timothée Lesort, Eugene Belilovsky, and Irina Rish.
\newblock Simple and {Scalable} {Strategies} to {Continually} {Pre}-train
  {Large} {Language} {Models}, March 2024.
\newblock URL \url{http://arxiv.org/abs/2403.08763}.
\newblock arXiv:2403.08763 [cs].

\bibitem[Karpukhin et~al.(2020)Karpukhin, Oğuz, Min, Lewis, Wu, Edunov, Chen,
  and Yih]{karpukhin_dense_2020}
Vladimir Karpukhin, Barlas Oğuz, Sewon Min, Patrick Lewis, Ledell Wu, Sergey
  Edunov, Danqi Chen, and Wen-tau Yih.
\newblock Dense {Passage} {Retrieval} for {Open}-{Domain} {Question}
  {Answering}, September 2020.
\newblock URL \url{http://arxiv.org/abs/2004.04906}.
\newblock arXiv:2004.04906 [cs].

\bibitem[Radford et~al.(2021)Radford, Kim, Hallacy, Ramesh, Goh, Agarwal,
  Sastry, Askell, Mishkin, Clark, Krueger, and
  Sutskever]{radford_learning_2021}
Alec Radford, Jong~Wook Kim, Chris Hallacy, Aditya Ramesh, Gabriel Goh,
  Sandhini Agarwal, Girish Sastry, Amanda Askell, Pamela Mishkin, Jack Clark,
  Gretchen Krueger, and Ilya Sutskever.
\newblock Learning {Transferable} {Visual} {Models} {From} {Natural} {Language}
  {Supervision}, February 2021.
\newblock URL \url{http://arxiv.org/abs/2103.00020}.
\newblock arXiv:2103.00020 [cs].

\bibitem[Hendrycks et~al.(2021{\natexlab{a}})Hendrycks, Burns, Basart, Critch,
  Li, Song, and Steinhardt]{hendrycks2021ethics}
Dan Hendrycks, Collin Burns, Steven Basart, Andrew Critch, Jerry Li, Dawn Song,
  and Jacob Steinhardt.
\newblock Aligning ai with shared human values.
\newblock \emph{Proceedings of the International Conference on Learning
  Representations (ICLR)}, 2021{\natexlab{a}}.

\bibitem[Hendrycks et~al.(2021{\natexlab{b}})Hendrycks, Burns, Basart, Zou,
  Mazeika, Song, and Steinhardt]{hendryckstest2021}
Dan Hendrycks, Collin Burns, Steven Basart, Andy Zou, Mantas Mazeika, Dawn
  Song, and Jacob Steinhardt.
\newblock Measuring massive multitask language understanding.
\newblock \emph{Proceedings of the International Conference on Learning
  Representations (ICLR)}, 2021{\natexlab{b}}.

\bibitem[Guo et~al.(2024)Guo, Fu, Zhang, Zhao, and
  Shen]{guo2024efficientcontinualpretrainingmitigating}
Yiduo Guo, Jie Fu, Huishuai Zhang, Dongyan Zhao, and Yikang Shen.
\newblock Efficient continual pre-training by mitigating the stability gap,
  2024.
\newblock URL \url{https://arxiv.org/abs/2406.14833}.

\end{thebibliography}
\bibliographystyle{unsrtnat}

%%%%%%%%%%%%%%%%%%%%%%%%%%%%%%%%%%%%%%%%%%%%%%%%%%%%%%%%%%%%%%%%%%%%%%%%%%%%%%%
%%%%%%%%%%%%%%%%%%%%%%%%%%%%%%%%%%%%%%%%%%%%%%%%%%%%%%%%%%%%%%%%%%%%%%%%%%%%%%%
% APPENDIX
%%%%%%%%%%%%%%%%%%%%%%%%%%%%%%%%%%%%%%%%%%%%%%%%%%%%%%%%%%%%%%%%%%%%%%%%%%%%%%%
%%%%%%%%%%%%%%%%%%%%%%%%%%%%%%%%%%%%%%%%%%%%%%%%%%%%%%%%%%%%%%%%%%%%%%%%%%%%%%%
\newpage
\appendix
% \onecolumn
\section{MTEB Subtasks used for Ablation Study}\label{appendix:15_tasks}
To save computation, we performed our ablation study based on $15$ sub-tasks of MTEB, following \cite{behnamghader2024Llm2vec}. The definition of $15$ subtasks are:
\begin{itemize}
    \item ArguAna
    \item ClimateFEVER
    \item NFCorpus
    \item SciFact
    \item StackOverflowDupQuestions
    \item SciDocsRR
    \item BiorxivClusteringS2S
    \item MedrxivClusteringS2S
    \item TwentyNewsgroupsClustering
    \item SprintDuplicateQuestions
    \item Banking77Classification
    \item EmotionClassification
    \item MassiveIntentClassification
    \item STS17
    \item SICK-R
    \item STSBenchmark
\end{itemize}

\section{Examples Language Generation with Special Token(s)}
Table \ref{tab_example} shows examples of compressing input text to one special token then recovering it using the GEM Llama 3-8B model. In this experiment, we simply inserted one special token after the input text, then save kv cache (the key and value tensor of every layer) of the special token and finally run auto-regressive decoding using kv cache with empty input. 

The table indicates with as few as one special token, the model is able to compress text and fully recover it for input up to $\approx30$ tokens. However, as the input getting longer, the model struggles to recover the identical text but still captures the key semantics. Adding more special tokens could handle longer inputs and this would be studied in our future works.

\begin{table*}[t]
\caption{Examples of compressing the input text (left column) to one special token and then decode the text from the special token (right column) using the GEM Llama 3-8B model. We highlighted the differences between input text and recovered text.}
\label{tab_example}
\begin{center}
\begin{small}
% \begin{sc}
\begin{tabular}{p{8cm}|p{8cm}}
\toprule
Input text & Recovered Text \\
\midrule
Paula is boring to climb down all steps.&Paula is boring to climb down all steps. \\\midrule

% The worn shawls aren't concealed. & The worn shawls aren't concealed. \\\midrule
Can Veronica ever heal Melissa? & Can Veronica ever heal Melissa? \\\midrule
% Large decoder-only language models (LLMs) have achieved remarkable success&Large decoder-only language models (LLMs) have achieved remarkable success\\\midrule

Large decoder-only language models (LLMs) have achieved remarkable success in \textcolor{orange}{generation and reasoning} tasks&Large decoder-only language models (LLMs) have achieved remarkable success in \textcolor{orange}{language reasoning and generation} tasks \\\midrule

Large decoder-only language models (LLMs) have achieved \textcolor{orange}{remarkable} success in \textcolor{orange}{generation and reasoning} tasks, where they generate \textcolor{orange}{text responses given instructions}. & Large decoder-only language models (LLMs) have achieved \textcolor{orange}{great} success in \textcolor{orange}{reasoning and generation} tasks, where they \textcolor{orange}{provide language outputs} generate\textcolor{orange}{d from choice questions} \\\midrule

Text encoding models play a crucial role in compressing long text sequences into semantic representations, facilitating downstream tasks like information retrieval, clustering, and classification. & Text encoding models play a crucial role in compressing long text sequences into semantic representations, facilitating downstream tasks like information retrieval, clustering, and classification. \\\midrule
Canoeing: Then, the men pivot the canoe away from the bow paddlers while paddling fast the oars. After, the men slides the canoe forward using the oars to advance. Next , the men flip the canoe and stops at the bow paddlers. &  Canoeing: Then, the men pivot the canoe away from the bow paddlers while paddling fast the oars. After, the men slides the canoe forward using the oars to advance. Next , the men flip the canoe and stops at the bow paddlers. \\\midrule

Large \textcolor{orange}{decoder-only} language \textcolor{orange}{models (LLMs)} have achieved \textcolor{orange}{remarkable} success \textcolor{orange}{in generation and reasoning tasks, where they generate text responses given instructions}. However, many \textcolor{orange}{applications, e.g., retrieval augmented generation (RAG), still rely on separate embedding models to generate text embeddings, which can complicate the system and introduce discrepancies in understanding of the query between the embedding model and LLMs.} & Large\textcolor{orange}{Decoding }language\textcolor{orange}{s (legen languages)} have achieved \textcolor{orange}{great} success \textcolor{orange}{and ability to generate relevant information,} however, many \textcolor{orange}{generation tasks where addressed solely using language generators, which form the basis of reinforcement learning, still require other types of decoder, e.g., subroutines, to increase the decoding accuracy and can introduce complications in the system of generating reference sentences over the language-based decoder and thus, make}\\
\bottomrule
\end{tabular}
% \end{sc}
\end{small}
\end{center}
\vskip -0.1in
\end{table*}

% \subsection{Examples of Query Results}
% \todo{add query examples here}
% \section{Example of Text Generation}
% \todo{add some examples of question answering from the finetuned model. Please refer to Table 2 http://arxiv.org/abs/2307.06945 as an examples.}
%%%%%%%%%%%%%%%%%%%%%%%%%%%%%%%%%%%%%%%%%%%%%%%%%%%%%%%%%%%%%%%%%%%%%%%%%%%%%%%
%%%%%%%%%%%%%%%%%%%%%%%%%%%%%%%%%%%%%%%%%%%%%%%%%%%%%%%%%%%%%%%%%%%%%%%%%%%%%%%

\end{document}